\documentclass[10pt,twocolumn,letterpaper]{article}

\usepackage{cvpr}
\usepackage{times}
\usepackage{epsfig}
\usepackage{graphicx}
\usepackage{amsmath}
\usepackage{amssymb}
\usepackage{mathrsfs}
\usepackage[numbers]{natbib}
\usepackage[usenames,dvipsnames,svgnames,table]{xcolor}
\usepackage{booktabs}
\usepackage{soul}
\usepackage{xspace}
\usepackage{algorithm}
\usepackage{subcaption}

\usepackage[pagebackref=true,breaklinks=true,letterpaper=true,colorlinks,bookmarks=false]{hyperref}

\graphicspath{{images/}}

\newcommand{\comment}[1]{}

\newcommand{\amosrmk}[1]{{\color{blue}{\bf as: #1}}}

\newcommand{\davideM}[1]{{\color{purple} #1}}
\newcommand{\davideMrmk}[1]{{\color{purple}{\bf jm: #1}}}

\newcommand{\Eb}{Event-based\ }
\newcommand{\eb}{event-based\ }

\newcommand{\fb}{frame-based\ }

\cvprfinalcopy % *** Uncomment this line for the final submission

\def\cvprPaperID{***} % *** Enter the CVPR Paper ID here

\ifcvprfinal\fi
\begin{document}

%%%%%%%%% TITLE
\title{A Large Scale Event-based Detection Dataset for Automotive}

%% TODO fix authors
%% git clone https://asironi@bitbucket.org/asironi/cvpr_19_eb_feature.git

\author{
	Pierre de Tournemire \quad %\\
	%{\tt\small pdetournemire@prophesee.ai}\\
	%PROPHESEE, Paris, France
	%\and
	Davide Nitti \quad %\\
	%{\tt\small dnitti@prophesee.ai}\\
	%PROPHESEE, Paris, France
	%\and
	Etienne Perot \quad
	%{\tt\small eperot@prophesee.ai}\\
	%PROPHESEE, Paris, France
	%\and
	Davide Migliore \quad 
	%\vspace{2mm}
	%{\tt\small dmigliore@prophesee.ai}\\
	%PROPHESEE, Paris, France
	Amos Sironi
	\vspace{1mm}\\
	{\tt\small \{pdetournemire, dnitti, eperot, dmigliore, asironi\}@prophesee.ai}\vspace{1mm}\\
	%{\tt\small dmigliore@prophesee.ai asironi@prophesee.ai}\\
	%{\tt\small \{ pdetournemire@prophesee.ai dnitti@prophesee.ai eperot@prophesee.ai}\\
	%	{\tt\small dmigliore@prophesee.ai asironi@prophesee.ai}\\
	PROPHESEE, Paris, France\thanks{This work was supported in part by the EU H2020 ULPEC project.}
}

\maketitle
%\thispagestyle{empty}

% %%%%%%%%% ABSTRACT
\begin{abstract}
								
	We introduce the first very large detection dataset for event cameras. 
	The dataset is composed of more than 39 hours of automotive recordings 
	acquired with a 304x240 GEN1 sensor.
	\comment{\davideM{we should say that the dataset it recorded in France?}.
	\amosrmk{we say it in the main text, I think it's enough}}
	It contains open roads and very diverse driving scenarios, ranging from urban, 
	highway, suburbs and countryside scenes, as well as different weather and 
	illumination conditions.  

	Manual bounding box annotations of cars and pedestrians contained in the recordings 
	are also provided at a frequency between 1 and 4Hz, 
	yielding more than 255,000 labels in total.
	We believe that the availability of a labeled dataset of this size will contribute 
	to major advances in event-based vision tasks such as object detection 
	and classification. We also expect benefits in other tasks such as 
	optical flow, structure from motion and tracking, where for example, 
	the large amount of data can be leveraged by self-supervised learning methods.
	\comment{\davideM{we do not provide gt for this right? In my opinion better not to say}
	\amosrmk{for tracking you can still use the same labels. For other tasks, 
	self-supervised learning can be used. I try to make it more exlicit}}
\end{abstract}

% % %%%%%%%%% BODY TEXT
\section{Introduction}
\label{sec:intro}		

Large datasets are a fundamental ingredient for modern computer vision~\cite{Deng09,Lin14}.
On one side, the availability of large benchmarked datasets allowed
objective and common evaluation of novel algorithms against the state-of-the-art
~\cite{Everingham10,Krizhevsky12,Lin14}. 
The diverse and large amount of samples in these datasets guarantee 
robustness in real-world applications, compared to small datasets.
On another side, large labeled datasets opened the possibility to train very 
deep machine learning models~\cite{Krizhevsky12,Lecun15,He16}, able 
to generalize well also on samples drawn from different
distributions than the train set.  

\begin{figure}
	\centering
	\begin{tabular}{@{}c@{ }c@{}}%
		\includegraphics[width=0.5\linewidth]{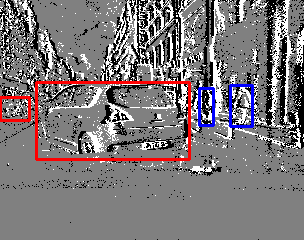} & %       
		\includegraphics[width=0.5\linewidth]{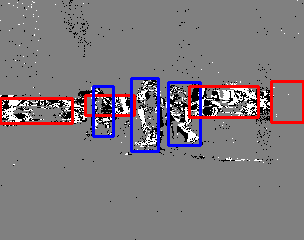} \\
		\includegraphics[width=0.5\linewidth]{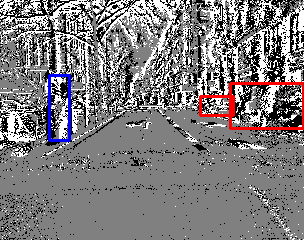} & %       
		\includegraphics[width=0.5\linewidth]{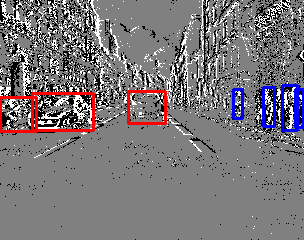}		
	\end{tabular}
	\caption{Examples from the ATIS Automotive Detection Dataset. We release more than 39
		hours of automotive recordings from an \eb ATIS sensor~\cite{Posch11}, 
		together with more than 255,000 manual bounding box annotations of cars and pedestrians.
		In the figure, events are shown by cumulating 100ms of data into a binary histogram, where 
		white corresponds to positive contrasts changes and black to negative ones. 
		Blue bounding boxes correspond to pedestrian labels, red bounding boxes to cars.
		To the best of our knowledge, the ATIS Automotive Detection Dataset is the largest 
		\eb dataset ever released.
	}
	\label{fig:intro}
\end{figure}

Event-based vision, which is the field of performing visual tasks from the 
output of an event camera~\cite{Gallego19},
is a much younger research filed compared to standard frame-based computer vision.
Event cameras~\cite{Lichtsteiner08,Posch11,Serrano13} are a recent sensor representing visual information in the form 
of an asynchronous stream of $\{(x,y,p,t)\}$ events, representing
log-luminosity contrast changes at time $t$ and location $(x,y)$. With 
$p$ a binary variable indicating the sign of the contrast change, Fig.~\ref{fig:intro}.

Event cameras are characterized by very high dynamic range ($>$120dB),
extremely high temporal resolution (in the order of microseconds)
and adaptive data rate (in fact, events are produced only at the time and positions of a contrast change). 
As a consequence, event cameras do not suffer from oversampling, undersampling and motion blur.
%I changed the "of" with a "from" 

Similarly to frame-based vision, low-level \eb vision tasks
such as noise filtering~\cite{Khodamoradi18}, 
edge detection~\cite{Lagorce15a,Lee19}, clustering~\cite{Barranco18}, etc. 
have been addressed using analytical and geometrical methods. 
	
However, as the complexity of the task increases, the number of variables 
and parameters of a system aiming at solving them also increases.
Tuning this large number of parameters without a data-driven approach becomes
soon impractical.
	
For this reason, event-based vision is increasingly adopting machine learning
techniques~\cite{Zhu18,Maqueda18,Manderscheid19,Rebecq19}.
Together with these methods, several datasets have been released~\cite{Orchard15,Sironi18,Zhu18b,Bi19}.
	
However, the size of these datasets is much smaller compared to their frame-based counterparts. 
To give an example, the largest labeled \eb dataset to date for classification~\cite{Bi19} is 
composed of 100,800 samples, while Imagenet~\cite{Deng09} contains 14 millions  
labeled images!

Due to the scarce availability of real event-based dataset, many researchers
turned to simulator-based solutions~\cite{Rebecq18,Gehrig19}. This approach is appealing  
because it simplifies label generation and it can be complementary to real data collection.
However, real sequences remain fundamental in order to 
capture the unique properties of the \eb sensors, 
which can not be obtained starting from sequences of frames,
and to be robust to noise and unidealities, which are hard to simulate 
with an idealized model.

With this work, we release more than 39 hours of automotive recordings taken with an 
GEN1~\cite{Posch11} event camera in realistic driving conditions. 
Each recording has a variable duration between tens of minutes 
and several hours. We also collect and release 
228,123 cars and 27,658 pedestrians bounding boxes,
obtained by manually labeling the gray-level images provided by the GEN1 sensor,
at a frequency of 1Hz, 2Hz or 4Hz, depending on the sequence. 
	
To the best of our knowledge, this is the largest event-based dataset ever released
\comment{\davideM{automotive} \amosrmk{this is the largest in absolute, not restricted to automotive.
We can specify that it is automotive related somewhere else if you want.}}
in terms of both total number of hours and total number of labels.
It is also the only automotive one providing accurate bounding box localization for 
a multi-class detection task.
	
Thanks to this dataset, we reduce the gap between \fb datasets and \eb datasets.
In this way, we hope that also the gap in accuracy between \fb and \eb vision systems 
will sharply decrease.
We expect benefits for both supervised tasks, such as detection and classification,
and self-supervised ones, such as optical flow, monocular-depth estimation as well as tracking.

\section{Related Work}
\label{sec:related_work}		

In this section, we describe the main existing \eb datasets. 
We start by describing labeled datasets for recognition and classification tasks,
and then we describe datasets generated for other tasks, 
such as visual odometry and optical flow. 

\paragraph{\Eb Datasets for Recognition}
Early \eb datasets have been generated by converting existing 
\fb datasets to an event representation~\cite{Orchard15,Serrano15,Hu16}.
For example in~\cite{Orchard15}, the MMINST~\cite{Lecun98} and Caltech-101~\cite{Fei06} 
datasets have been converted to events by moving an event camera in front of a screen
displaying the images.
Similarly in~\cite{Hu16} the~\cite{Kristan15,Griffin07,Reddy13} 
\fb datasets have been converted by positioning a static
\eb camera in front of a monitor playing the datasets.

The advantage of these approaches is that it is possible to create large datasets 
without the need of costly manual labeling.
The drawback is that the planar display and the limited frequency 
of the screen results in unnatural and very constrained event sequences.

Because of this, recent works have focused in realizing real world datasets for recognition.

For example, in~\cite{Sironi18} 12,336 car examples were 
manually labeled and extracted from open road driving recordings, 
together with 11,693 background samples.

In~\cite{Amir17} and~\cite{Bi19} instead, two gesture recognition datasets were built 
by asking several human subjects to perform the gestures in front of the camera.
For example,~\cite{Bi19} contains 100,800 examples and it is the largest classification 
dataset available to date in terms of number of labels. 
However, each sample contains only 100ms of data, cropped from longer sequences.
This reduces the actual variability contained in the training data and amounts to
less than 3 hours of data.

The authors of~\cite{Miao19} acquired several recordings from an event camera
to build 3 datasets for surveillance applications: one for 
pedestrian detection, one for an action recognition and one for fall detection.
The labels for the pedestrian dataset were obtained by building
image-like representation from 20ms of events and then manually annotating them.
This is the first event-based dataset from real data for detection. 
However, the dataset is composed by only 12 sequences of 30 seconds.

Finally, the authors of~\cite{Calabrese19} collect an \eb datasets for 3D pose estimation.
Ground-truth was obtained using motion capture and infrared cameras together 
with reflective markers positioned on the human subjects joints.

\paragraph{\Eb Datasets for Visual Odometry, Optical Flow and Stereo}

Other datasets focus on different applications than recognition, and they can leverage
complementary sensors or techniques for automated labeling.

In~\cite{Binas17}, 12 hours of driving sequences are obtained during day and night time.
Various car information, such as vehicle speed, GPS position, driver steering angle, 
are associated to the dataset. 
The dataset has been used for end-to-end steering angle prediction~\cite{Maqueda18}
and also to generate pseudo-labels for event data, by running standard \fb detectors 
on the graylevel images provided with the dataset~\cite{Chen18}.

The authors of~\cite{Zhu18b}, collected sequences using several complementary 
sensors coupled to the event camera. In particular, 
depth ground truth is provided thanks to the use of a lidar. 
This dataset has been extended to obtain optical flow ground-truth~\cite{Zhu18}.

In~\cite{Leung18} 10 hours of stereo recordings have been acquired together 
with pose ground-truth at 100Hz.
In~\cite{Mitrokhin19} a motion segmentation dataset is realized, while 
the autors of~\cite{Manderscheid19} focus instead of the problem of 
corner detection by realizing a dataset in the same spirit of the \fb~\cite{Mikolajczyk05}. 
Finally, it is worth mentioning the first color \eb dataset~\cite{Scheerlinck19}.

An \eb simulator is available in~\cite{Rebecq18} to generate event sequences from standard videos.
For example, it has been used in~\cite{Rebecq19} for learning 
to reconstruct a graylevel images from events. 
And in~\cite{Gehrig19} together with a slowmotion frame-based method
to convert \fb datasets to \eb ones. 

Using a simulator to convert \fb data into \eb
is a valid and complementary approach to real data collection. However,
the need of real data is still essential to fully leverage properties
of the event cameras, such as high-dynamic range and high temporal resolution, 
which are not properly captured by standard \fb cameras. 
Moreover, accurately replicating noise, sensor unidealities, read-out effects, etc.
of real \eb cameras can be challenging using an idealized simulation model.

The amount of datasets released in the past years confirms the growing interest 
in \eb vision and a very active community. However, the size and the annotations
of the available datasets is still very minor compared to \fb datasets such as 
Imagenet~\cite{Deng09} or COCO~\cite{Lin14}. Yet, accurate annotations and very large datasets are 
critical for designing and evaluating vision systems that can operate reliably in realworld 
situations.

In the next section, we describe the first detection \eb dataset with accurate manual annotation of 
cars and pedestrians in real driving conditions. The datasets contains more that 39 hours of data,
and it is the largest \eb dataset ever made available to the public.

\section{The ATIS Automotive Detection Dataset}
\label{sec:dataset}		

%eb cameras 
\subsection{Event Cameras}
\label{subsec:event_cameras}	

Event cameras are a relatively recent type of sensor 
encoding visual information in the form of asynchronous events~\cite{Lichtsteiner08,Posch11,Serrano13}. 
An event corresponds to a change in the log-luminosity intensity at a given pixel location.

In an event camera, the photosensitive part is composed by a 2D array of independent pixels. 
Whenever a pixel detects a change in illuminance intensity, it emits
an event containing its $(x,y)$ position in the pixel array, the
microsecond timestamp $t$ of the observed change and its polarity $p$. 
The polarity encodes whether the illuminance intensity
increased ($p=1$) or decreased ($p=0$).

Compared to standard frame cameras, event cameras have higher temporal resolution, 
higher dynamic range and lower power consumption.
Thanks to these characteristics, event cameras find many applications
in automotive, robotics and IoT, where low latency, 
robustness to challenging lighting conditions and power 
consumption are critical requirements.

Many event cameras are currently available in the market~\cite{Psee19,Son17,Guo17,Brandli14}. 
Some of them also provide graylevel information in forms of 
synchronous frames~\cite{Brandli14,Guo17}
or by asynchronous event-based measurements~\cite{Psee19}.

In this work, we consider a Gen1 304x240 camera~\cite{Posch11}.
The luminous intensity measures from the camera were used 
to generate standard gray-level images at a given frequency.
The images were then manually annotated by human subjects to  
generate ground-truth bounding boxes around objects of interest.
The labeling procedure is explained in detail in Sec.~\ref{subsec:labeling}.
 
% driving instructions and main scenarios
\subsection{Data Collection}
\label{subsec:data}
A GEN1 camera was mounted behind the windshield of a car and connected to 
a laptop for data recording. 
Different human drivers, independent from the authors, 
were asked to perform several rides in different scenarios,
but always driving naturally. There are minor variations in the camera position due
to repeated mountings of the camera.

The scenarios include city with dense traffic, city with low traffic, highway, countryside,
small villages and suburbs. All recordings were done on France roads, 
mainly in, but not limited to, the Ile-de-France region. Recordings duration varies from
tens of minutes to a maximum of several consecutive hours.
\begin{figure*}[htpb]
	\centering
	\begin{tabular}{@{}c@{ }c@{ }c@{ }c@{}}{\hspace{0mm}}%
		% %
		%\multicolumn{4}{c}{\hspace{0mm}%
		%,height=0.9\linewidth
		\includegraphics[width=0.25\linewidth]{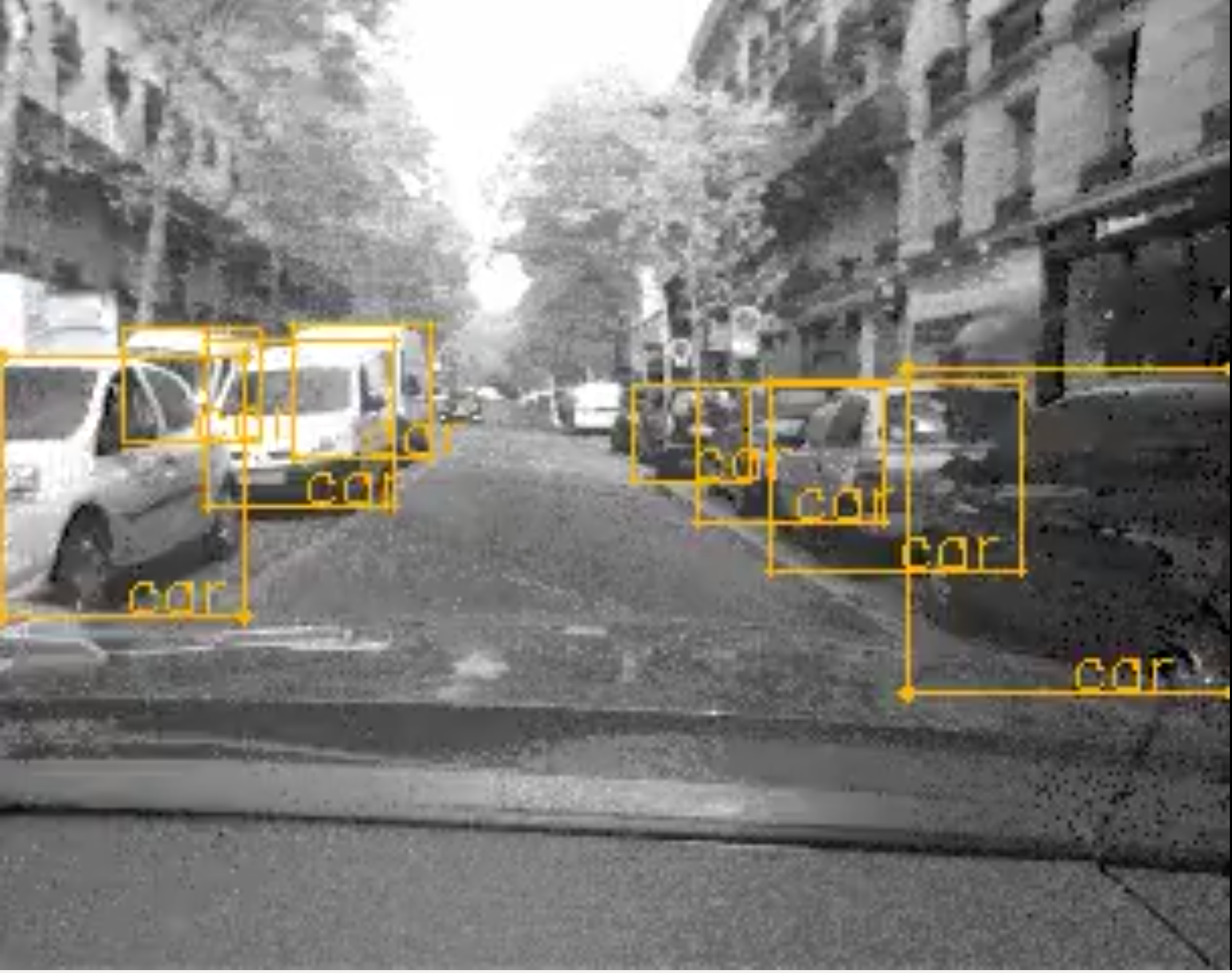} & %    
		\includegraphics[width=0.25\linewidth]{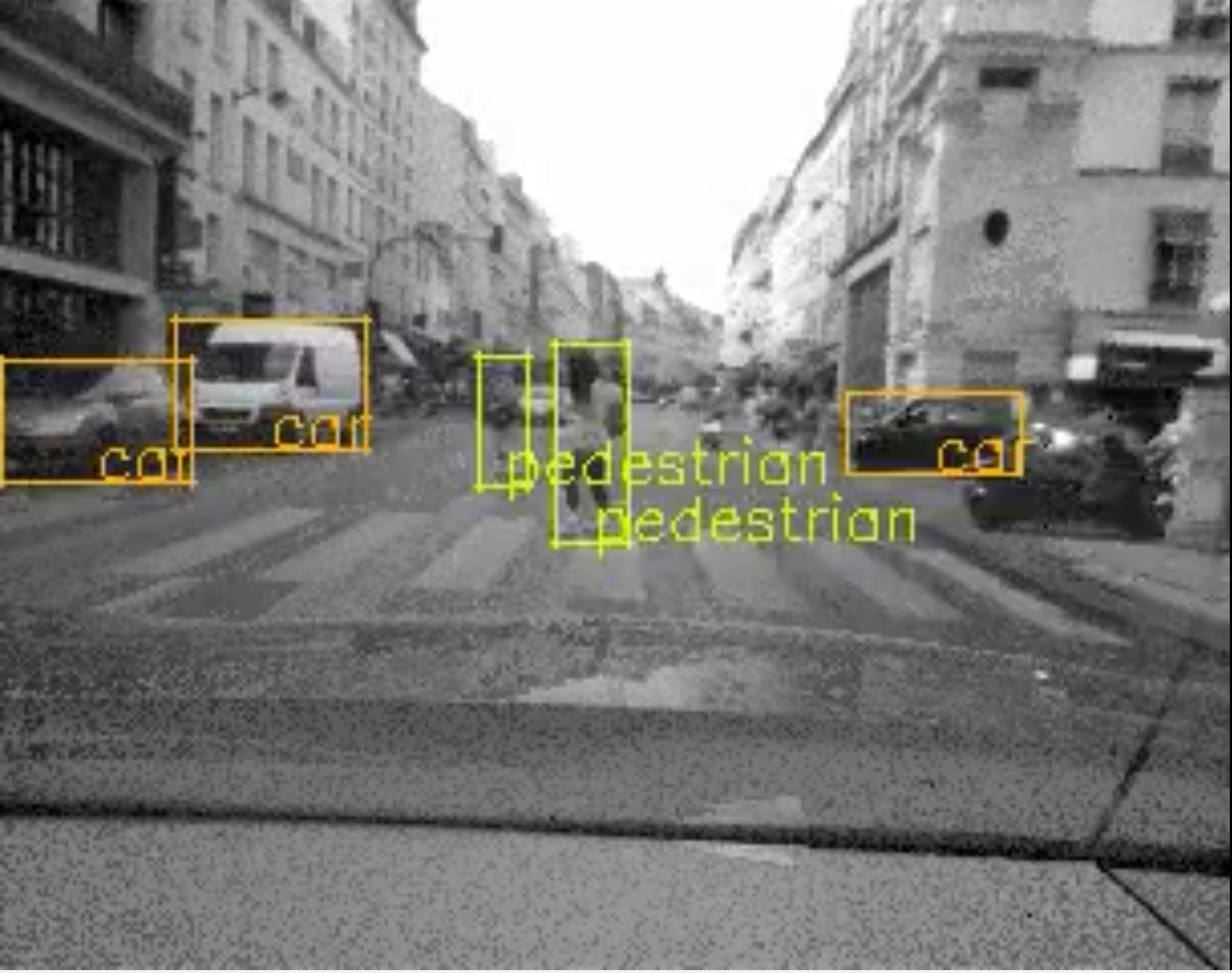} & %    
		\includegraphics[width=0.25\linewidth]{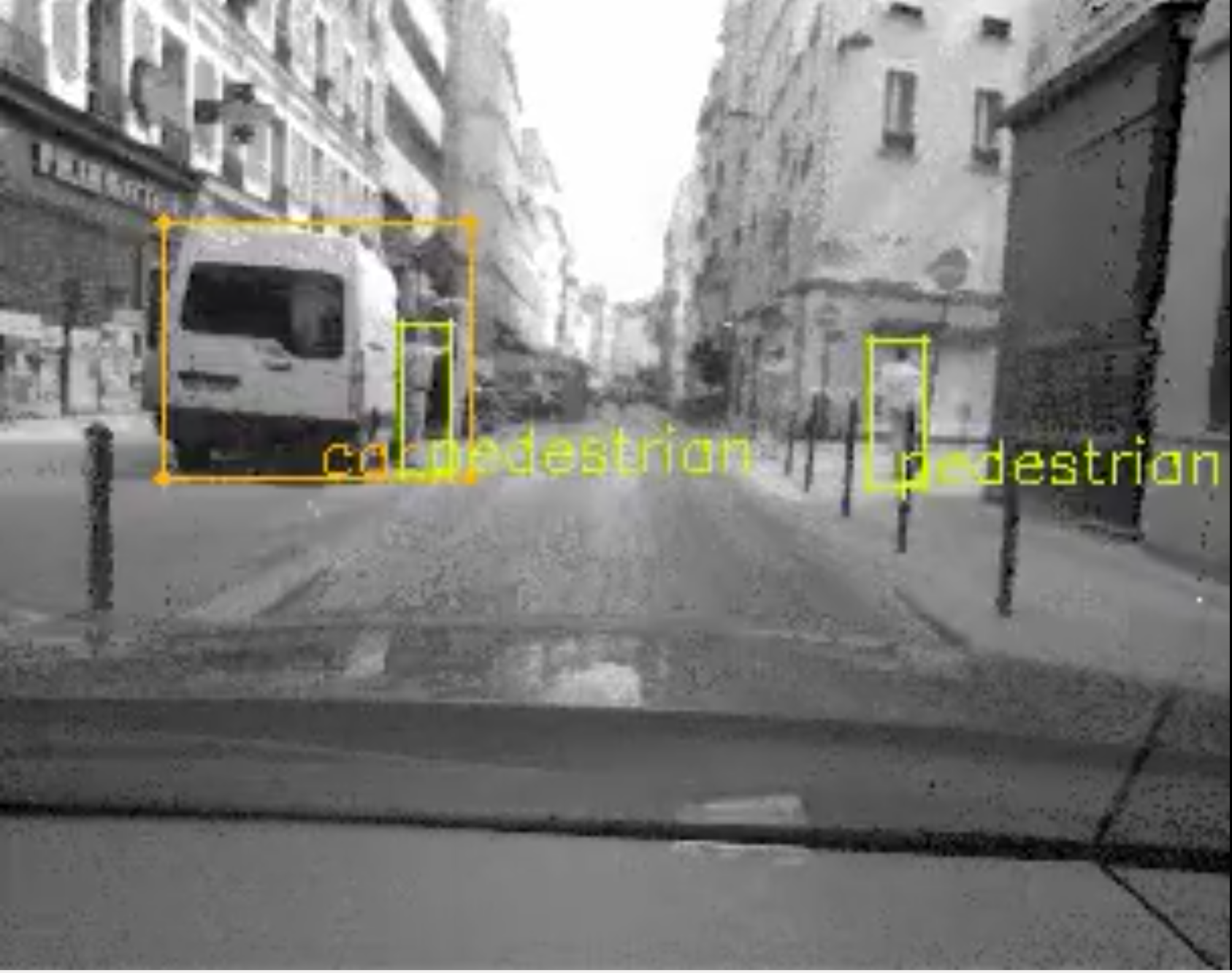} & %    
		\includegraphics[width=0.25\linewidth]{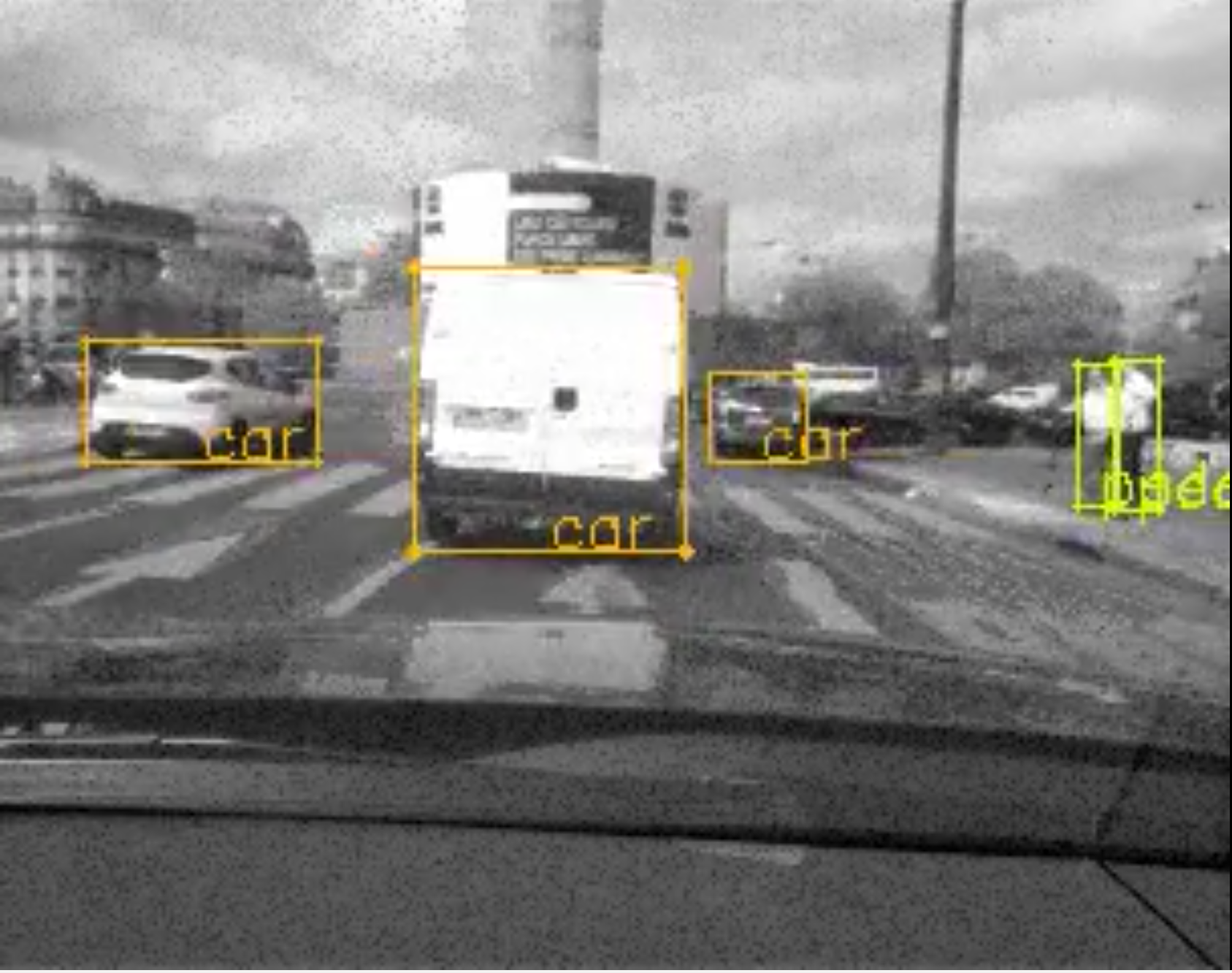}
		%}%
		%{\hspace{1.0cm} \bf (a) Events} & {  \hspace{0.2cm}  {\bf (b) FAST~\cite{Mueggler17}} } & {\hspace{0cm}  \bf (c) Ours}   % \\
		% \multicolumn{2}{l}{\hspace{-1mm}\includegraphics[width=0.9\linewidth]{intro_c1.pdf}}%
		%
		%\includegraphics[height=0.5\linewidth,width=0.3\linewidth]{introv1.pdf}&%
		%{%
		% \setlength{\fboxsep}{0pt}%
		% \setlength{\fboxrule}{1pt}%
		% \fboxrule=0.05pt%
		%  \fbox{%
		%
		%}
		%}
		%
	\end{tabular}
	% \vspace{0.1em}
	\caption{Examples of gray-level images used for manual annotation, with overlaid 
		bounding boxes as drawn by the human labelers. The images were generated by aggregating
		the asynchronous intensity measurements of the ATIS sensors. 
		Images and the events stream share the same pixel array, 
		moreover, to each image is associated the 
		precise timestamp of the last event used to generate them. 
		Thanks to these properties of the ATIS sensor, the bounding boxes can
	be directly used as ground truth for the event stream.}
	% \vspace{-3mm}
	\label{fig:graylabels}
\end{figure*}

The data collection campaign was conducted over an entire year, from March 2017 to March 2018,
and at variable times of the day, assuring a large variety of lightening and weather conditions.  
\comment{\davideM{There are differences of sesons? weather condition?}
\amosrmk{yes, as mentioned in the text.}}
A total of 39.32 hours, split among 121 recordings were collected, resulting in
about 750GB of uncompressed raw event data. 
For comparison, a gray-scale \fb camera working 
at the same resolution and acquiring at a frequency of 120fps
(i.e. 100 times lower temporal resolution compared to the event camera),
would generate more than 1.2TB of data
\footnote{Ignoring compression and assuming 1 byte per pixel}.

In the next section, we describe how the data were manually annotated.

%labeling process and instructions
\subsection{Labeling Protocol}
\label{subsec:labeling}
The GEN1 sensor provides along with the change detection events, also
gray-level measurements.
These measurements can be used to build a gray-level image at any desired time.
The time of the last measurement used to generate them is associated to the image,
providing images with the same temporal resolution as the event stream.
Moreover, since gray-level images and the events stream share the same pixels array, 
annotations on the images can directly be used as ground truth for the event stream, 
without the need of any calibration or rectification step.

Since our primary goal is object detection, we favor low-frequency annotations 
in order to maximize the variety of objects aspects and scenes. Because of this,
we generate images at a 1, 2 or 4Hz. These images were then 
given to human annotators to draw bounding boxes around cars and pedestrians.
% Instructions given to the annotators are listed below.

A detailed set of instruction has been provided to the annotators to reduce 
ambiguity and discrepancies between annotations.

Due to the resolution and image quality of GEN1 images,
objects of size smaller than 30 pixels have been discarded.
Concerning occlusions, an object is annotated if it is visible for more than 75\%. 
In which case, the bounding box is drawn on the whole extend of the object.

Buses, trucks, and large vehicles are not considered as cars and therefore
have not been annotated. Similarly, for motorbikes and two-wheelers.
People moving on skateboards or kick-scooters have been labeled as pedestrians, 
while people sitting inside cars or in buildings have been ignored.

After annotation, we obtained a total of 228,123 cars and 27,658 pedestrians bounding boxes.
More statistics about the datasets are given in Sec.~\ref{sec:statistics}.
Example graylevel images together with manual annotations are shown in Fig.~\ref{fig:graylabels}.

%format and download link
\subsection{Dataset Format and Download}
\label{subsec:format}
We split the recordings into train, validation and test sets.
To avoid overlap between train and test splits, 
each single recording session is the same split.  

In order to facilitate the training of deep learning methods, we cut the continuous
recordings into 60 seconds chunks.
This yield to a total of 2359 samples: 1460 for train, 470 for test and 429 for validation.

Each sample is provided in a binary .dat format,
where events are encoded using 4 bytes 
for the timestamps and 4 bytes for the position and the polarity.
More precisely, 14 bits are used for the $x$ position, 14 bits for the $y$ position 
and 1 bit for the polarity. Gray-level measurements are not provided with the dataset.

Bounding box annotations are provided in a numpy format.
Each numpy array contains the following fields:
\begin{itemize}
	\item \texttt{ts}, timestamp of the box in microseconds
	\item \texttt{x}, abscissa of the top left corner in pixels
	\item \texttt{y}, ordinate of the top left corner in pixels
	\item \texttt{w}, width of the boxes in pixel
	\item \texttt{h}, height of the boxes in pixel
	\item \texttt{class\_id}, class of the object: 0 for cars and 1 for pedestrians	      	          
\end{itemize}

We make the obtained dataset publicly available through the following link 
\href{https://www.prophesee.ai/2020/01/24/prophesee-gen1-automotive-detection-dataset/
}{\texttt{https://www.prophesee.ai/2020/01/24/\\prophesee-gen1-automotive-detection-dataset/}}.
We also provide a sample code together with the dataset to load and visualize 
some samples from the dataset with the corresponding annotations.

For evaluating the accuracy of a detection method, 
we consider the same metrics used for the COCO dataset~\cite{Lin14}.
Together with the released code, we provide a wrapper and 
an example on how to apply the evaluation metrics on our dataset.

\section{Analysis and Statistics}
\label{sec:statistics}		
In this section, we extract some statistics from the ATIS Automotive Detection Dataset
and we compare it to existing \eb datasets.

\begin{figure}
	\centering
	\includegraphics[trim=20 20 30 20, clip, width=1\linewidth]{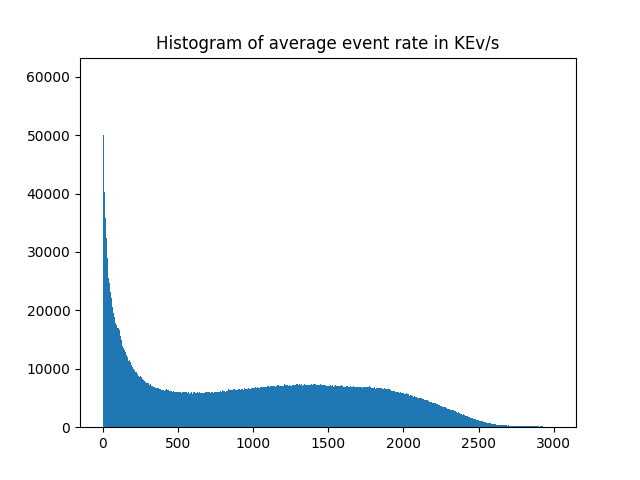}	
	\caption{Histogram of the average event rate computed over $1ms$ time windows
		for the entire ATIS Automotive Detection Dataset. Most of the recordings 
		have event rate below $200Kev/s$, but higher rates are also present, 
		with peaks up to $3Mev/s$. 
	Notice that the event rate was computed without any filtering of the data.}
	% \vspace{-3mm}
	\label{fig:event_rate}
\end{figure}

% events stats
We start by analyzing the properties of raw events stream. 
In particular, we study the rate of event stream generated during the recordings.
In order to do this, we split the recordings in $1ms$ intervals and compute 
the average event rate in this interval, without any filtering or noise removal. 
We then build an histogram from these
measurements. As we can see from Fig.~\ref{fig:event_rate}, the majority 
of the samples contain very low data rate, below $200Kev/s$. 
However, the distribution has a long tail, with maximum peaks reaching up to $3Mev/s$.
These peaks corresponds to scene with very strong lightening changes, such as
flickering lights or fast repeated transitions from bright sun to shadow. 

\begin{figure}
	\centering
	\begin{tabular}{@{}c@{ }c@{}}%
		\includegraphics[width=0.5\linewidth]{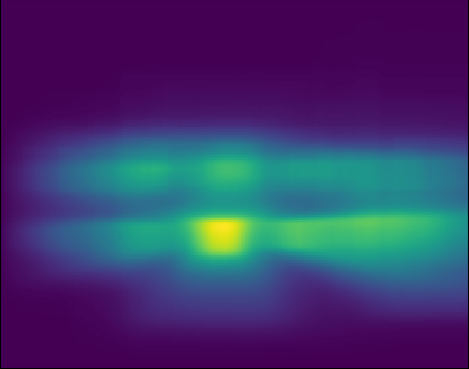} & %                                                          
		\includegraphics[width=0.5\linewidth]{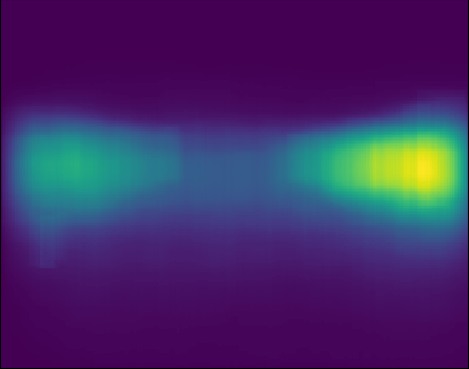}	\\
		{\hspace{0cm} \bf (a)}                                 & {  \hspace{0cm}  \bf (b) }%	  
	\end{tabular}
	\caption{Heatmaps of the manually labeled bounding boxes in dataset, 
		for \textbf{(a)} cars and \textbf{(b)} pedestrians. Heatmaps have been computed 
	by counting for each pixel the number of boxes covering that pixel.}
	\label{fig:heatmaps}
\end{figure}

% labels stats
We then study the distribution of the annotated bounding boxes.
Similarly to~\cite{Dollar12}, we compute the heat map on the location
of the bounding box (Fig.~\ref{fig:heatmaps}). For cars, we observe two 
principal horizontal axis, corresponding to two main positioning 
of the camera inside the car. This is less visible in the pedestrian heatmap,
probably because pedestrians are mostly seen in city recordings, 
where the camera position was most stable. 
We also notice a larger number of boxes in the right part of the image. 
This is due to the fact that driving is mostly conducted on the right lane of the road 
and therefore objects on the left part appear smaller and are more often discarded by the 
30 pixel diagonal threshold.

\begin{figure*}
	\centering
	\begin{tabular}{@{}c@{ }c@{ }c@{ }c@{}}{}%
		%trim=left bottom right top,
		\includegraphics[trim=24 20 24 20, clip, width=0.25\linewidth]{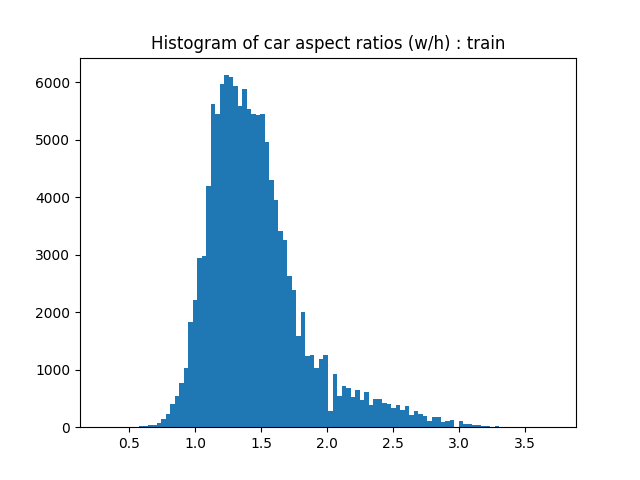} &
		\includegraphics[trim=24 20 24 20, clip, width=0.25\linewidth]{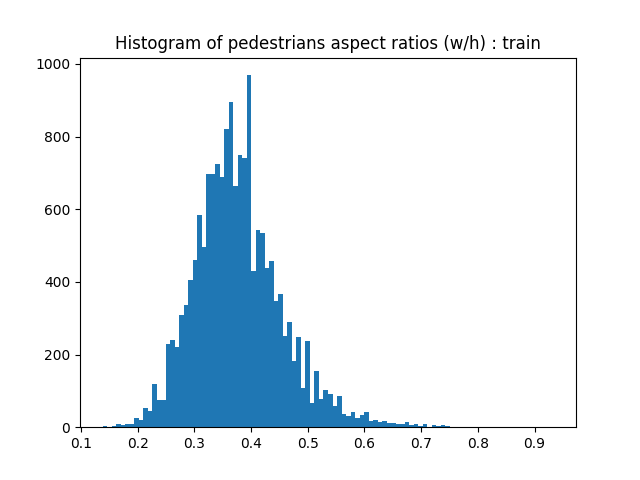} &
		\includegraphics[trim=24 20 24 20, clip, width=0.25\linewidth]{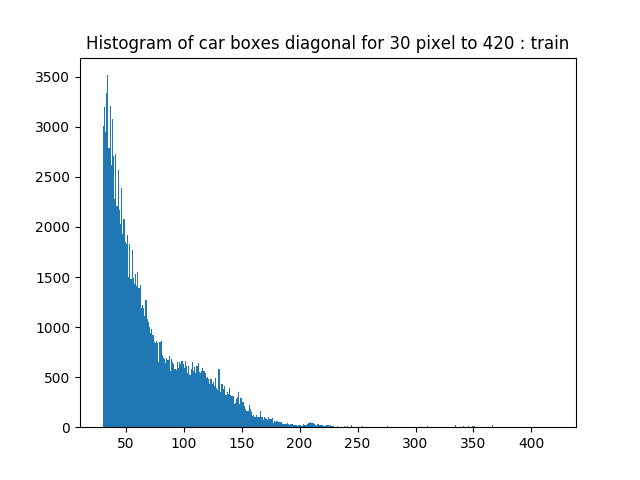} &
		\includegraphics[trim=24 20 24 20, clip, width=0.25\linewidth]{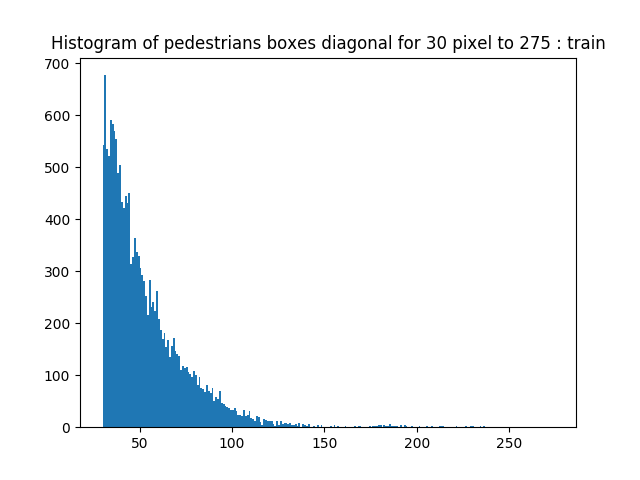} \\
		\includegraphics[trim=24 20 24 20, clip, width=0.25\linewidth]{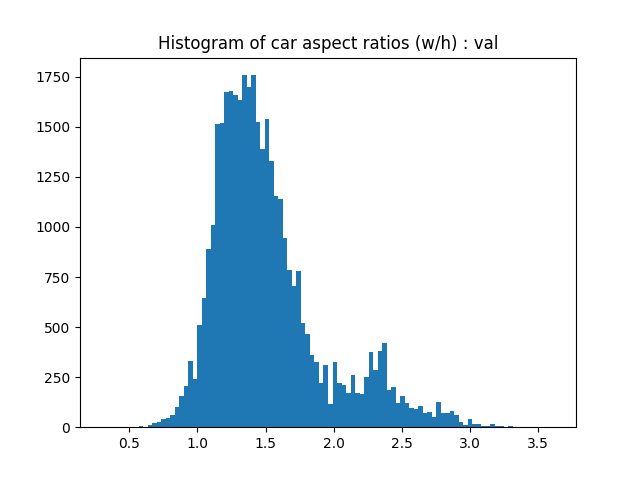} &
		\includegraphics[trim=24 20 24 20, clip, width=0.25\linewidth]{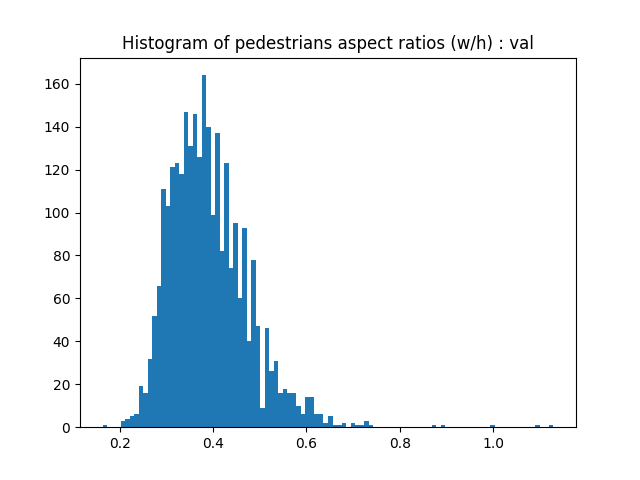} &
		\includegraphics[trim=24 20 24 20, clip, width=0.25\linewidth]{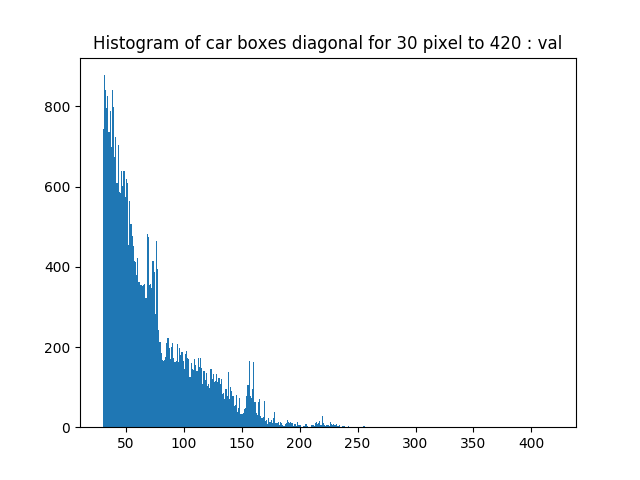} &
		\includegraphics[trim=24 20 24 20, clip, width=0.25\linewidth]{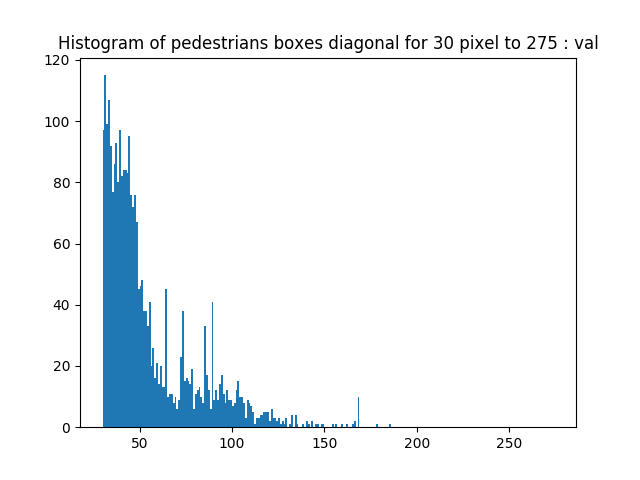} \\
		\includegraphics[trim=24 20 24 20, clip, width=0.25\linewidth]{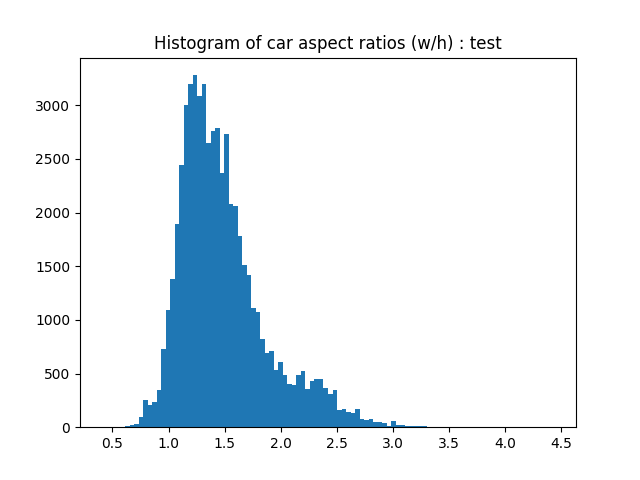} &
		\includegraphics[trim=24 20 24 20, clip, width=0.25\linewidth]{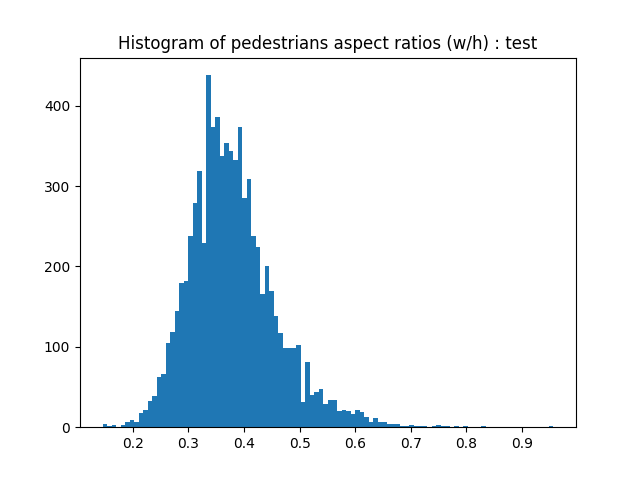} &
		\includegraphics[trim=24 20 24 20, clip, width=0.25\linewidth]{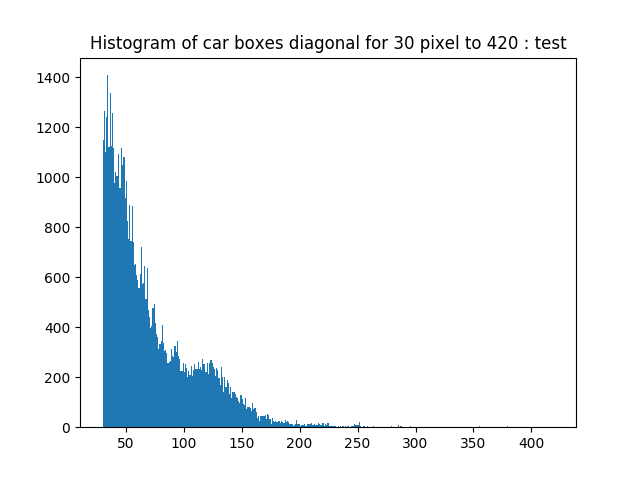} &
		\includegraphics[trim=24 20 24 20, clip, width=0.25\linewidth]{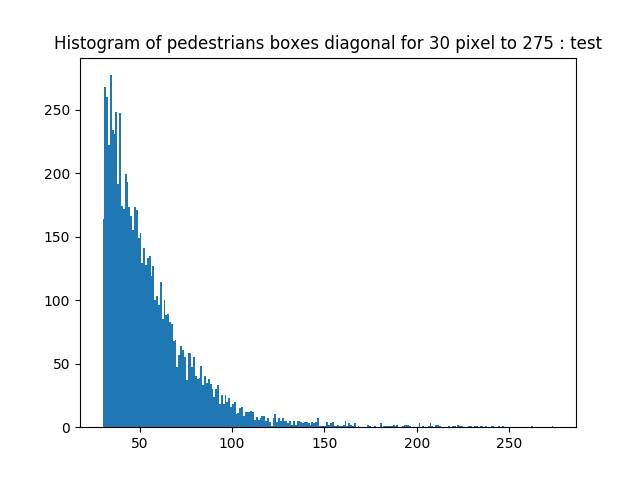} \\
		{ \bf (a)} & { \bf (b) } %
		  & { \bf (c) } & { \bf (d) } 
	\end{tabular}
	\caption{Statistics of the manually labeled bounding boxes in the dataset.
		{\bf (a,b)} Histogram of aspect ratio (width over height) of the bounding boxes for
		cars and pedestrians respectively. {\bf (c,d)} Histogram of the bounding boxes diagonal 
		for cars and pedestrians respectively. Bounding boxes of objects with diagonal smaller than 30 pixels %
		have not been annotated. The three rows correspond to train, validation and test splits respectively.
	We observe a similar distribution in the three splits.}
	\label{fig:box_stats}
\end{figure*}

In Fig.~\ref{fig:box_stats}\textbf{(a,b)} we show the histogram of the bounding box aspect ratio,
computed as width over height.
Histograms are computed on the train, validation and test splits independently. 
For pedestrians, the aspect ratio has a gaussian distribution,
with mean around 0.35; while for cars, the histogram
is closer to a two-modal distribution. This is due to fact that the aspect ratio 
varies depending on the point of view of the car: 
cars seen from the front or from behind have ratio closer to 1,
while cars seen from the side have larger aspect ratio.

In Fig.~\ref{fig:box_stats}\textbf{(c,d)}, we show instead the histogram of the bounding box diagonal.
For both cars and pedestrian we observe a long tail distribution, starting from 
the 30 pixel threshold set for manual annotation.
Finally, we observe that train, validation and test splits, 
have similar statistics. 

%table/graph of comparison with other datasets
Finally, we compare the dataset with other existing \eb dataset.
As shown by Tab.~\ref{tab:datasets}, the GEN1 Automotive Detection Dataset is 
3 time larger than the DDD17~\cite{Binas17} dataset in terms of hours and 
has about 22 times more labels than the~\cite{Miao19} pedestrian dataset.
In terms of number of labels, the~\cite{Bi19} dataset is the second largest one,
with approximately 2.5 less labels than ours. 
However,~\cite{Bi19} considers a classification task and each sample is only 100 ms long. 

\begin{table*}[htpb]
	\caption{Comparison of available \eb datasets for different tasks. The GEN1 Automotive Detection (GAD)
		Dataset is the largest in terms of both number of hours and number of manual annotations. It is also the only 
	automotive dataset with semantic bounding box labels for detection.}
	\begin{center}
		\begin{tabular}{@{}llcccc@{}}
			\toprule
			\textbf{Dataset}                   & \textbf{Task}              & \textbf{Max Sample Time (s)} & \textbf{Total Time (h)} & \textbf{\# Labels} & \textbf{\# Classes} \\% & \textbf{Real World} \\
			\hline
			{AAD Dataset (this work)}          & Detection for Automotive   & 60 (10,020$^*$)  & 39.32                & 255,781            & 2                   \\%    & yes                 \\
			{Pedestrian Dataset~\cite{Miao19}} & Detection for Surveillance & 30                        & 0.10                 & 11,667             & 1                   \\%     & yes                  \\
			\hline
			{N-Mnist~\cite{Orchard15}}         & Object Classification      & 0.3                       & 5.83                 & 70,000             & 10                  \\%   & no                  \\
			{N-Caltech101~\cite{Orchard15}}    & Object Classification      & 0.3                       & 0.76                 & 9,146              & 101                 \\% & no                  \\
			{N-Cars~\cite{Sironi18}}           & Object Classification      & 0.1                       & 0.68                 & 24,029             & 2                   \\%  & yes                 \\
			{DVS-Gestures~\cite{Amir17}}       & Gesture Recognition        & 6                        & 2.24                 & 1,342              & 11                  \\%  & yes                 \\
			{ASL-DVS~\cite{Bi19}}              & Gesture Recognition        & 0.1                       & 2.80                 & 100,800            & 24                  \\%  & yes                 \\
			\hline
			{MVSEC~\cite{Zhu18b}}              & Stereo, Flow, VO           & 1,500                     & 1.13                 & -                  & -                   \\%   & yes                 \\
			{DDD17~\cite{Binas17}}             & Autonomous Driving         & 3,135                     & 12                   & -                  & -                   \\%  & yes                 \\
			\bottomrule
			\multicolumn{6}{l}{\small $^*$ Samples are obtained by splitting continuous recordings into 60s chunks.
				The longest of the original recordings is 10,020s long.}
		\end{tabular}%}
	\end{center}
	\label{tab:datasets}
	%\vspace{1ex}
%     {\raggedright $^*$ This is where authors provide additional information about the data, including whatever notes are needed. \par}
\end{table*}
%\footnotemark
%\footnotetext{Samples are obtained by splitting continuous recordings into 60 s chunks.
%The longest of the original recordings is 10,020 s long.}

\section{Conclusion}
\label{sec:conclusion}		
We presented the first large automotive dataset for detection with event cameras. 
Thanks to this dataset, we open the way to the training of deep learning 
models for detection on event-based cameras.
We also expect benefits in other applications, such as object tracking, 
and unsupervised learning of optical flow
and monocular depth, among others.

We hope that the \eb research community will \comment{\davideM{I don't like this greatly}
\amosrmk{what about considerably? but then we use considerable below} \davideMrmk{I suggest: We are sure the research community will have great benefits from this dataset}}
greatly benefit from this dataset
and that it will soon become a reference benchmark. We also believe 
that thanks to the availability of such a large dataset, 
the accuracy of event-based vision systems will undergo considerable advances.

% %%%%%%%%% BIBLIOGRAPHY

{\small
	\bibliographystyle{ieee}
	\bibliography{short,biblio}
}

\end{document}